\documentclass[10pt, conference, compsocconf]{IEEEtran}
\usepackage[pdftex]{graphicx}
\ifCLASSINFOpdf
\else
\fi
\usepackage{amsmath,graphicx}

\begin{document}
\topmargin=0mm
%
% paper title
% can use linebreaks \\ within to get better formatting as desired
\title{Fusion of Deep and Non-Deep Methods for Fast Super-Resolution of Satellite Images}

% author names and affiliations
% use a multiple column layout for up to two different
% affiliations

\author{
\IEEEauthorblockN{Gaurav Kumar Nayak, Saksham Jain, R Venkatesh Babu, Anirban Chakraborty}
\IEEEauthorblockA{Department of Computational and Data Sciences\\
Indian Institute of Science, Bangalore, India\\
Email: \{gauravnayak, sakshamjain, venky, anirban\}@iisc.ac.in}
}
\maketitle

\begin{abstract}
In the emerging commercial space industry there is a drastic increase in access to low cost satellite imagery. The price for satellite images depends on the sensor quality and revisit rate. This work proposes to bridge the gap between image quality and the price by improving the image quality via super-resolution (SR). Recently, a number of deep SR techniques have been proposed to enhance satellite images. However, none of these methods utilize the region-level context information, giving equal importance to each region in the image. This, along with the fact that most state-of-the-art SR methods are complex and cumbersome deep models, the time taken to process very large satellite images can be impractically high. We, propose to handle this challenge by designing an SR framework that analyzes the regional information content on each patch of the low-resolution image and judiciously chooses to use more computationally complex deep models to super-resolve more structure-rich regions on the image, while using less resource-intensive non-deep methods on non-salient regions. Through extensive experiments on a large satellite image, we show substantial decrease in inference time while achieving similar performance to that of existing deep SR methods over several evaluation measures like PSNR, MSE and SSIM.

\end{abstract}


% For peer review papers, you can put extra information on the cover
% page as needed:
% \ifCLASSOPTIONpeerreview
% \begin{center} \bfseries EDICS Category: 3-BBND \end{center}
% \fi
%
% For peerreview papers, this IEEEtran command inserts a page break and
% creates the second title. It will be ignored for other modes.
\IEEEpeerreviewmaketitle

\section{Introduction}
% no \IEEEPARstart
Satellites images are used to observe the earth's surface from outer space in order to provide rich visual information. Such information is often useful in many important real-world applications, such as natural disaster warning \cite{bredemeyer2018radar}, exploration of resource \cite{li2016hyperspectral}, land cover classification \cite{jiang2018superpca, he2018remote, fang2018hyperspectral, zhu2018deformable}, weather/environment monitoring and many more. However, often the quality and resolution of these satellite images are affected due to the limitations in the imaging equipment, the space to ground station communication bandwidth, transmission noise etc. \cite{lu2019satellite}. The acquired low resolution (LR) images, resulting from the aforementioned hardware limitations, may not be suitable for the downstream analytics tasks such as object detection and fine grained classification, precision mapping and measurement etc. To overcome such challenges, super-resolution (SR) techniques are used towards improving the spatial resolution of the collected images \cite{tsai1984multiframe, lim2009super, lu2019satellite, wang2018esrgan}.

Deep learning has been successfully applied to various vision tasks including object recognition, detection, action recognition, image/video analytics and processing, optical flow estimation, image captioning, etc. Deep learning approach is a data dependent approach, where a large amount of image data along with the ground truth information is used for training a neural network with many parameters. The design of the network architecture plays a crucial role in obtaining better results. Many SR techniques that use deep learning have been proposed for general/natural images. For example, SRCNN \cite{dong2015image} performs end-to-end learning from low resolution to high-resolution images via fully convolutional network. The method uses a bi-cubic interpolation as its pre-processing step followed by the extraction of overlapping patches, via convolution, as high dimensional vectors with as many feature maps as their dimensions. The vectors are
then non-linearly mapped to each other and subsequently aggregated in the form of patches to get the reconstructed SR image that is supposed to be as close to the ground truth as possible. Another method, VDSR \cite{kim2016accurate} uses a very deep convolution network inspired by VGG-net used for ImageNet classification. In this approach, the residue is predicted from the input low resolution image and the final SR image is obtained by adding the residue with the interpolated low resolution image. In EPSCN \cite{shi2016real}, the final convolutional layer predicts $r^2$ channels corresponding to each pixel of low resolution image. This approach is simple and quite powerful for super-resolving.
% You must have at least 2 lines in the paragraph with the drop letter
% (should never be an issue)

\begin{figure*}[htb]
	
	\begin{minipage}[b]{1.0\linewidth}
		\centering
		\centerline{\includegraphics[width=0.9\linewidth, height =0.4\linewidth] {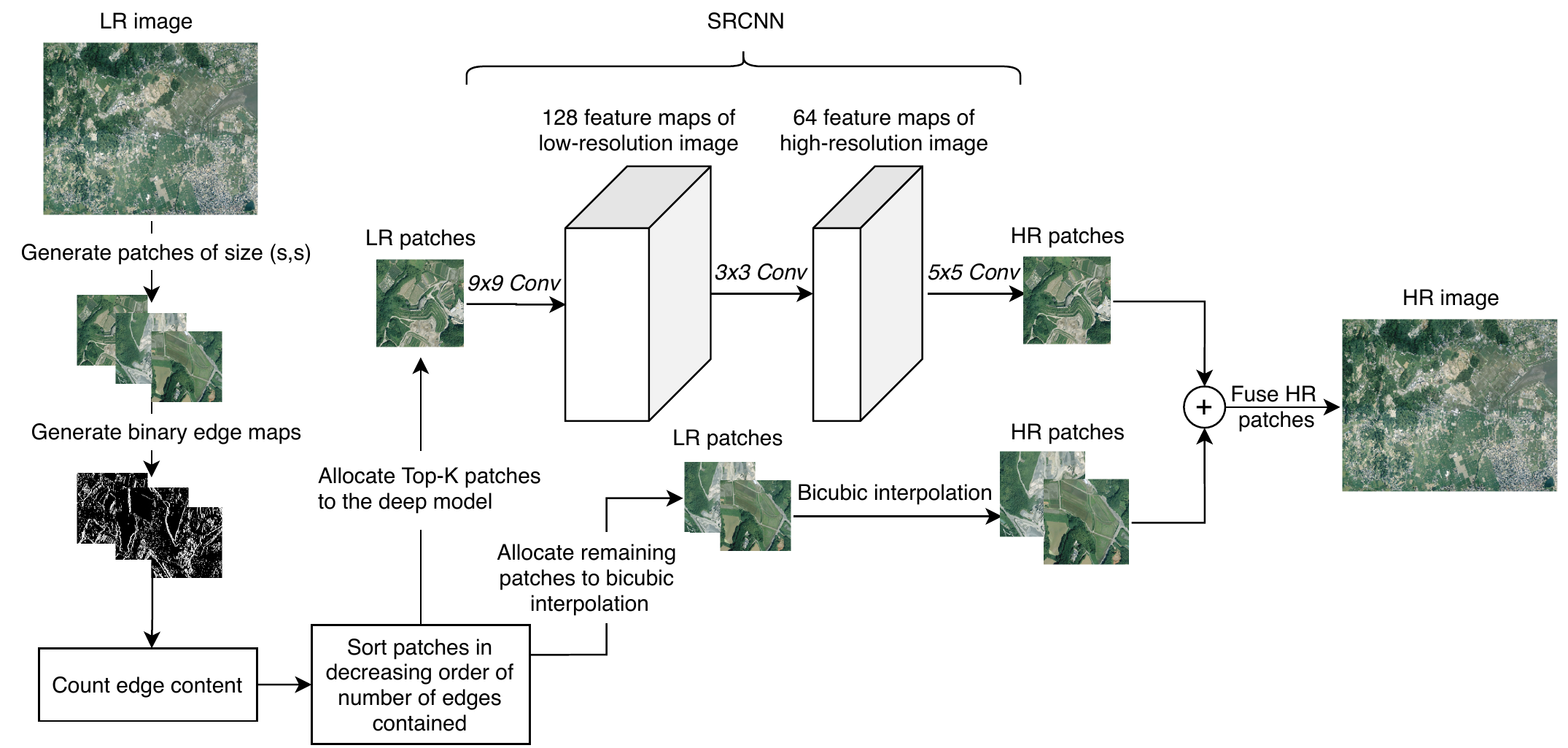}}

		%\centerline{(a) Result 1}\medskip
	\end{minipage}
	\caption{Proposed Method: The low-resolution image is divided into patches, and edge maps are generated where the top $k$ patches (descending order) based on edge content are allocated to the deep model, and the remaining patches are allocated to bicubic interpolation. Finally the high-resolution patches obtained from the deep and non-deep methods are fused to obtain the output image} 
	\label{fig:network}
	
\end{figure*}

Most of the aforementioned super-resolution approaches do not rely on context information and treat all regions equally, whereas in satellite images, the context plays a crucial role in determining the objects, which usually occupy a tiny region (for example, the object floating on a large water body is most likely a ship or boat). Since the satellite images are huge in size, it is a good candidate for super-resolving only for salient object regions. This would ensure that the images are processed quickly despite their large size. Leveraging this, we propose a novel approach where we utilize the deep SR models only on high frequency regions which are more informative whereas fast non-deep methods like Bicubic interpolation are used to super-resolve on the low frequency regions.

The rest of the paper is organized as follows:  section 2 list out the major contributions of this work, section 3 discusses the proposed framework in detail, section 4 demonstrates the empirical evaluation and discussion on the results, and section 5 finally concludes the paper.

\section{Detailed Contributions}
\label{sec:pagestyle}

\begin{enumerate}
	\item To the best of our knowledge, this is the first work in the direction of selective deep  super-resolution on satellite and aerial imagery via intelligent fusion of deep and non-deep methods .
	\item We design a simple and intuitive selection criteria, based on edge information for deciding image regions which are more salient and thus better served by super-resolution via deep models, and non-salient regions to be super-resolved using fast non-deep methods.
	\item The proposed approach is model-agnostic and can easily be coupled with any state-of-the-art deep super-resolution model. 
	\item Our methodology enables significantly faster super-resolution with marginal loss in performance as compared to the deep model. We demonstrate the veracity of our claims using SRCNN due to its simplicity and popularity.

\end{enumerate}

\section{Proposed Method}
\label{sec:typestyle}

\begin{figure*}[htb]
	\begin{minipage}[b]{1.0\linewidth}
		\centering
		\centerline{\includegraphics[width=0.8\linewidth, height=0.36\linewidth  ]{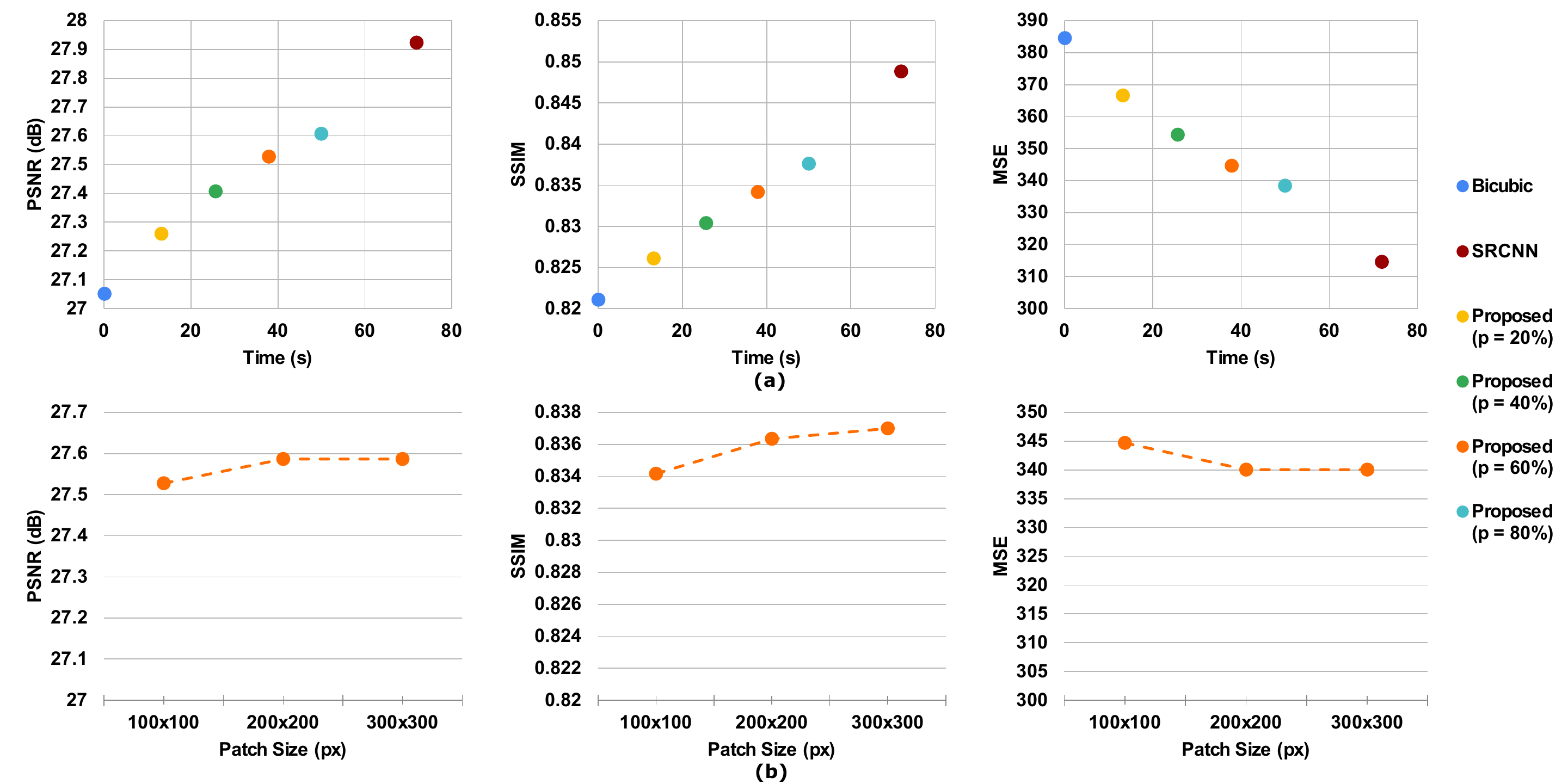}}
	\end{minipage}
	\caption{Ablation Study: (a) Scatter Plot describing the variation in the performance of Bicubic Interpolation, SRCNN and Proposed Method over time taken for super-resolution for a fixed patch size of $100 \times 100$ (b) The performance of the Proposed Method on different evaluation metrics over several choices of patch size.}
	\label{visualization}
	
\end{figure*}

%l from the publicly available
%codes provided by the authors, and all images are downsampled using the same bicubic kernel.
 The steps involved in our proposed method are diagrammatically depicted in Figure \ref{fig:network} and also described below in detail:
\begin{enumerate}
\item \textbf{Generating Patches:} The input low-resolution image is cropped into several non-overlapping patches of equal length ($s \times s$), which are then passed into the next module.
\item \textbf{Generating Binary Edge Map:} This module performs simple edge detection using gradients - first, the gradient vector at each pixel is computed by convolving the image with horizontal and
vertical derivative filters, and then gradient magnitudes are computed at each pixel. If magnitude at a pixel exceeds a threshold (taken here as 100 when the maximum pixel value is 255),
a possible edge point is reported. The module generates a binary edge map for each low-resolution image patch.
\item \textbf{Finding Amount of Edge Content:} This module simply counts the number of pixels where the edge point is reported, for each low-resolution image patch.
\item \textbf{Sorting:} The patches are sorted in decreasing order of the amount of edge content they contain.
\item \textbf{Allocating Patches:} This module allocates Top-K patches to the deep model for super-resolution while the remaining patches are simply upscaled using bicubic interpolation. Here, Top-K corresponds to the percentage of total number of low-resolution image patches allocated to the deep model.
\item \textbf{Fusing Patch Responses:} The final module collects the high-resolution image patches generated by bicubic interpolation and the deep model. Then these image patches are fused to obtain the final high-resolution image output. Here, the image reconstruction is analogous to completing a jigsaw puzzle, accomplished by simply placing the patches in the spatially correct order as in the original image.

The obtained high resolution image from the proposed method can be used for any downstream task.

\end{enumerate}

\vspace{-0.1 in}
\section{Experiments}
\label{sec:experiment}

In this section, we describe the experimental settings to evaluate our proposed approach. We also discuss about several ablations to decide the trade-off between speed and performance. Finally, we demonstrate the efficacy of the proposed method by achieving performance similar to deep model like SRCNN with significant drop in run time.
%across several metrics  %Thereby achieving fast super-resolution on satellite images. 

\subsection{Image Description} 
\label{subhead1}
To evaluate the performance of our proposed approach on the real-life application of super-resolution on satellite and aerial imagery, we use the aerial image of Kyushu University New Campus available on \cite{kyushu}. The original image, sized 12000 x 8486, was resized to 6000 x 4800 owing to hardware constraints, and treated as the high-resolution image. The low-resolution image is obtained as described in \cite{dong2015image} by downsampling high-resolution image by a factor of 2.

%\vspace{-0.1in}
\subsection{Pretrained Model}
\label{subhead2}
The SRCNN network  pretrained by following the same training paradigm defined in \cite{dong2015image} barring a few design changes, was taken as the choice of deep model for our proposed approach. Instead of the original network settings, i.e. filters $f_1 = 9, f_2 = 1, f_3 = 5$, and feature maps $n_1 = 64$ and $n_2 = 32$, the settings used were: $f_1 = 9, f_2 = 3, f_3 = 5, n_1 = 128$, and $n_2 = 64$. Further, the Adam optimizer was used with a learning rate of 0.0003 for all layers, and the MSE loss function was evaluated only by the difference between the central pixels of the input image and the network output. The training was done on 91 images for 200 epochs, with an upscaling factor of 2, and the convolutional layers were trained without padding in order to avoid any border effects. Although a fixed image size was used in training, the SRCNN input can be of arbitrary size at test time. We used the network weights from \cite{srcnn}.

\begin{figure*}[htb]
	\begin{minipage}[b]{1.0\linewidth}
		\centering
		\centerline{\includegraphics[width=0.8\linewidth, height=0.27\linewidth ]{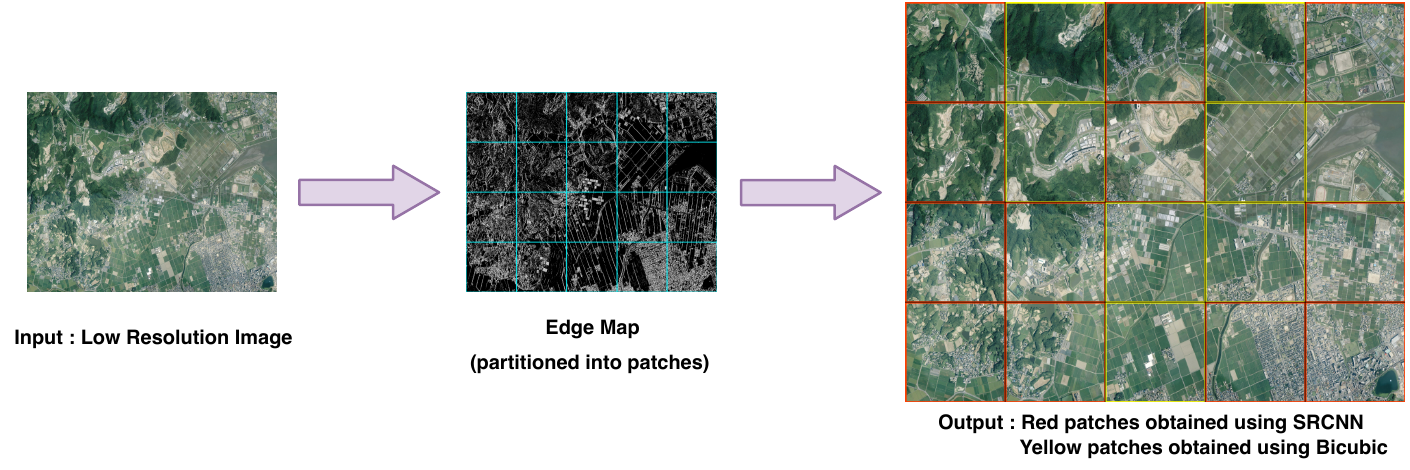}}
	\end{minipage}
	\caption{Demonstration of qualitative results obtained using our selective deep super-resolution approach on a given low resolution image.}
	\label{Qualitative Analysis}
	
\end{figure*}

\subsection{Evaluation Metrics}
\label{subhead5}
For evaluating our method, we use ``Mean-Squared Error" (MSE), ``Peak Signal-to-Noise Ratio" (PSNR), and ``Structural Similarity Index" (SSIM). We use the same evaluation protocol as mentioned in \cite{dong2015image} and the average results over 10 observations are reported. Given an image $(Y)$ of size $M \times N$ and a reference image $(X)$ of the same size, each pixel of X and Y is denoted by x and y respectively.

\textbf{Mean-Squared Error:} MSE measures the quality of Y by averaging the squared difference between its pixel values and that of X. It is always non-negative and lower values are better. A common problem with MSE is that it strongly depends on the intensity scaling of the image.
\begin{equation*}
    \operatorname{MSE}(X,Y)=\frac{1}{M N} \sum_{n=1}^{M} \sum_{m=1}^{N}[x\textsubscript{nm}-y\textsubscript{nm}]^{2}
    \tag{1}
\end{equation*}

\textbf{Peak Signal-to-Noise Ratio:} PSNR \cite{5596999} builds over the MSE by scaling it according to the image range. A higher PSNR value denotes a higher quality image, and vice versa. For 8-bit images:
\begin{equation*}
    \operatorname{PSNR}(X,Y)=10 \log _{10}\left(255^{2} / M S E(x, y)\right)
    \tag{2}
\end{equation*}

\textbf{Structural Similarity Index:} SSIM, developed by Wang et al.\cite{1284395}, is obtained by modelling an image distortion as a combination of three factors, namely, loss of correlation, contrast distortion, and luminance distortion. The commonly-used, specific form of SSIM is reduced to:
\begin{equation*}
    \operatorname{SSIM}(X,Y)=\frac{\left(2 \mu_{x} \mu_{y}+C_{1}\right)\left(2 \sigma_{x y}+C_{2}\right)}{\left(\mu_{x}^{2}+\mu_{y}^{2}+C_{1}\right)\left(\sigma_{x}^{2}+\sigma_{y}^{2}+C_{2}\right)}
    \tag{3}
\end{equation*}
where X and Y are two windows of common size $N \times N$, $\mu_{x}$ is the average of X, $\mu_{y}$ is the average of Y, $\sigma_{x}^{2}$ is the variance of X, $\sigma_{y}^{2}$ is the variance of Y, $\sigma_{x y}$ is the correlation coefficient of X and Y , $C_1 = k_1L^{2} $, $C_2 = k_2L^{2}$, $L$ is the dynamic range, and $k_1=0.01$ and $k_2=0.03$ by default.

\begin{table}[b]
    \centering
	\begin{tabular}{|c|c|c|c|c|}
		%\toprule
		\hline
		Method &
		{PSNR (dB)} & {SSIM} & {MSE} & {Time (s)}  \\
		\hline
		\hline
		
		%\midrule
		Bicubic & 27.05 & 0.82 & 384.62 &	0.10 \\
		\hline
		SRCNN & 27.92 & 0.85 & 314.65 &	71.91 \\
		\hline
		Proposed & 27.59 & 0.84 & 340.04 & 41.71\\
		\hline
		
		%\bottomrule
	\end{tabular}
\caption{Comparison of our proposed technique with a deep (SRCNN) and a non-deep (Bicubic) method}
\label{table1}
\end{table}

\subsection{Ablation Study}
\label{subhead6}

We do ablation on the choice of patch size ($s$), and percentage of patches super-resolved using the deep model ($p$) to investigate the trade-off between speed and performance. We tile the low-resolution image using patches of size $100 \times 100$, $200 \times 200$, and $300 \times 300$ at a time, while the hyperparameter for percentage of patches being super-resolved through deep model varies as \{$20$\%, $40$\%, $60$\%, $80$\%\}. The scatter plots in Fig. \ref{visualization} (a) clearly show that increasing $p$ significantly improves the performance. Particularly, the average PSNR values obtained by increasing $p$ for $s$ = 100 x 100 are 27.26 dB, 27.41 dB, 27.53 dB, and 27.61 dB respectively. However, the prediction speed also decreases with increase in $p$. For example, the average runtime values for the configurations are 13.22 s, 25.66 s, 37.92 s, and 50.00 s respectively. In this case, the best performance-speed tradeoff is obtained when 60\% of the low-resolution patches are super-resolved by the deep model. Hence, we set $p$ to 60\% in order to compare our method with other approaches.

Next, we examine the effect of enlarging the patch size to (i) 100 x 100, (ii) 200 x 200, and (iii) 300 x 300, keeping $p$ constant. As shown by the trend lines in Fig. \ref{visualization} (b), the performance of our model saturates beyond patch size of 200 x 200. Thus, we use this combination ($p=60\%$ and $s=200\times200$) as our hyperparameters for comparing the efficacy of our proposed method with Bicubic Interpolation and SRCNN. As observed from Table \ref{table1}, the PSNR, SSIM, and MSE values are 27.59 db, 0.84 and 340.04 respectively, and the time taken is 41.71 s.

%\vspace{-0.1in}
\subsection{Results and Discussion}
\label{subhead6}

Our proposed approach is qualitatively demonstrated in Figure \ref{Qualitative Analysis} where we perform selective deep super-resolution on patches of salient regions based on edge content, and bicubic interpolation for non-salient regions. It is evident from Table \ref{table1} that our proposed method shows a time advantage of 30.20 s, which is in line with an intuitive formula of \((1*t\textsubscript{srcnn} - (0.6*t\textsubscript{srcnn} + 0.4*t\textsubscript{bicubic}))\) s. On the other hand, it performs very close to the deep model on all evaluation metrics. Our choice of hyperparameters handles the tradeoff between speed and performance. Moreover, a suitable combination of $p$ and $s$, can help in prioritising either faster or better performance to obtain desired results depending upon the requirements. It is also worth noting that our proposed approach does not depend on the architecture of any deep SR model and can easily be used on top of it.

\section{Conclusion}
\label{sec:foot}
A key challenge in single image super-resolution for satellite and aerial imagery is inference time.
In this work, a novel approach has been proposed that exploits the power of deep methods and the speed of non-deep methods for the task in hand. Our framework, by design, greatly reduces the computational overhead and achieves performance close to the deep model across several evaluation measures. Our methodology also provides a flexible pipeline to decide the allocation of image patches to either deep or non-deep methods based on edge map which is a simple indicator of regional information content. More sophisticated indicators for identifying salient and informative regions in the satellite image will be explored in future work.% This is the first step towards selective super-resolution through intelligent fusion of deep and non-deep methods.
We hope that this work will be useful to the research community and generate further interest in the direction of selective deep super-resolution for satellite and aerial imagery.

% conference papers do not normally have an appendix

% use section* for acknowledgement

\section*{Acknowledgement}
This work was supported by an IISc-STC research project (Project code: ISTC/EDS/AC/423).

% trigger a \newpage just before the given reference
% number - used to balance the columns on the last page
% adjust value as needed - may need to be readjusted if
% the document is modified later
%\IEEEtriggeratref{8}
% The "triggered" command can be changed if desired:
%\IEEEtriggercmd{\enlargethispage{-5in}}

% references section

% can use a bibliography generated by BibTeX as a .bbl file
% BibTeX documentation can be easily obtained at:
% http://www.ctan.org/tex-archive/biblio/bibtex/contrib/doc/
% The IEEEtran BibTeX style support page is at:
% http://www.michaelshell.org/tex/ieeetran/bibtex/
%\bibliographystyle{IEEEtran}
% argument is your BibTeX string definitions and bibliography database(s)
%\bibliography{IEEEabrv,../bib/paper}
%
% <OR> manually copy in the resultant .bbl file
% set second argument of \begin to the number of references
% (used to reserve space for the reference number labels box)
%\begin{thebibliography}{1}

\iffalse
\bibitem{IEEEhowto:kopka}
H.~Kopka and P.~W. Daly, \emph{A Guide to \LaTeX}, 3rd~ed.\hskip 1em plus
  0.5em minus 0.4em\relax Harlow, England: Addison-Wesley, 1999.
 \fi

\bibliographystyle{IEEEtran}
\bibliography{IEEEabrv, ref}

% Generated by IEEEtran.bst, version: 1.14 (2015/08/26)
\begin{thebibliography}{10}
\providecommand{\url}[1]{#1}
\csname url@samestyle\endcsname
\providecommand{\newblock}{\relax}
\providecommand{\bibinfo}[2]{#2}
\providecommand{\BIBentrySTDinterwordspacing}{\spaceskip=0pt\relax}
\providecommand{\BIBentryALTinterwordstretchfactor}{4}
\providecommand{\BIBentryALTinterwordspacing}{\spaceskip=\fontdimen2\font plus
\BIBentryALTinterwordstretchfactor\fontdimen3\font minus
  \fontdimen4\font\relax}
\providecommand{\BIBforeignlanguage}[2]{{%
\expandafter\ifx\csname l@#1\endcsname\relax
\typeout{** WARNING: IEEEtran.bst: No hyphenation pattern has been}%
\typeout{** loaded for the language `#1'. Using the pattern for}%
\typeout{** the default language instead.}%
\else
\language=\csname l@#1\endcsname
\fi
#2}}
\providecommand{\BIBdecl}{\relax}
\BIBdecl

\bibitem{bredemeyer2018radar}
S.~Bredemeyer, F.-G. Ulmer, T.~H. Hansteen, and T.~R. Walter, ``Radar path
  delay effects in volcanic gas plumes: the case of l{\'a}scar volcano,
  northern chile,'' \emph{Remote Sensing}, vol.~10, no.~10, p. 1514, 2018.

\bibitem{li2016hyperspectral}
C.~Li, Y.~Ma, X.~Mei, C.~Liu, and J.~Ma, ``Hyperspectral unmixing with robust
  collaborative sparse regression,'' \emph{Remote Sensing}, vol.~8, no.~7, p.
  588, 2016.

\bibitem{jiang2018superpca}
J.~Jiang, J.~Ma, C.~Chen, Z.~Wang, Z.~Cai, and L.~Wang, ``Superpca: A
  superpixelwise pca approach for unsupervised feature extraction of
  hyperspectral imagery,'' \emph{IEEE Transactions on Geoscience and Remote
  Sensing}, vol.~56, no.~8, pp. 4581--4593, 2018.

\bibitem{he2018remote}
N.~He, L.~Fang, S.~Li, A.~Plaza, and J.~Plaza, ``Remote sensing scene
  classification using multilayer stacked covariance pooling,'' \emph{IEEE
  Transactions on Geoscience and Remote Sensing}, vol.~56, no.~12, pp.
  6899--6910, 2018.

\bibitem{fang2018hyperspectral}
L.~Fang, G.~Liu, S.~Li, P.~Ghamisi, and J.~A. Benediktsson, ``Hyperspectral
  image classification with squeeze multibias network,'' \emph{IEEE
  Transactions on Geoscience and Remote Sensing}, vol.~57, no.~3, pp.
  1291--1301, 2018.

\bibitem{zhu2018deformable}
J.~Zhu, L.~Fang, and P.~Ghamisi, ``Deformable convolutional neural networks for
  hyperspectral image classification,'' \emph{IEEE Geoscience and Remote
  Sensing Letters}, vol.~15, no.~8, pp. 1254--1258, 2018.

\bibitem{lu2019satellite}
T.~Lu, J.~Wang, Y.~Zhang, Z.~Wang, and J.~Jiang, ``Satellite image
  super-resolution via multi-scale residual deep neural network,'' \emph{Remote
  Sensing}, vol.~11, no.~13, p. 1588, 2019.

\bibitem{tsai1984multiframe}
R.~Tsai, ``Multiframe image restoration and registration,'' \emph{Advance
  Computer Visual and Image Processing}, vol.~1, pp. 317--339, 1984.

\bibitem{lim2009super}
K.~H. Lim and L.~K. Kwoh, ``Super-resolution for spot5-beyond supermode,'' in
  \emph{30th Asian Conference on Remote Sensing}, 2009, pp. 378--382.

\bibitem{wang2018esrgan}
X.~Wang, K.~Yu, S.~Wu, J.~Gu, Y.~Liu, C.~Dong, Y.~Qiao, and C.~Change~Loy,
  ``Esrgan: Enhanced super-resolution generative adversarial networks,'' in
  \emph{Proceedings of the European Conference on Computer Vision (ECCV)},
  2018, pp. 0--0.

\bibitem{dong2015image}
C.~Dong, C.~C. Loy, K.~He, and X.~Tang, ``Image super-resolution using deep
  convolutional networks,'' \emph{IEEE transactions on pattern analysis and
  machine intelligence}, vol.~38, no.~2, pp. 295--307, 2015.

\bibitem{kim2016accurate}
J.~Kim, J.~Kwon~Lee, and K.~Mu~Lee, ``Accurate image super-resolution using
  very deep convolutional networks,'' in \emph{Proceedings of the IEEE
  conference on computer vision and pattern recognition}, 2016, pp. 1646--1654.

\bibitem{shi2016real}
W.~Shi, J.~Caballero, F.~Husz{\'a}r, J.~Totz, A.~P. Aitken, R.~Bishop,
  D.~Rueckert, and Z.~Wang, ``Real-time single image and video super-resolution
  using an efficient sub-pixel convolutional neural network,'' in
  \emph{Proceedings of the IEEE conference on computer vision and pattern
  recognition}, 2016, pp. 1874--1883.

\bibitem{kyushu}
\BIBentryALTinterwordspacing
K.~University, ``{Photo Archives:} aerial photograph,'' 2010. [Online].
  Available:
  \url{http://suisin.jimu.kyushu-u.ac.jp/en/showcase/photo/aerial/h22/index.html}
\BIBentrySTDinterwordspacing

\bibitem{srcnn}
\BIBentryALTinterwordspacing
Y.~Maoke, ``Srcnn-keras,'' 2017. [Online]. Available:
  \url{https://github.com/MarkPrecursor/SRCNN-keras}
\BIBentrySTDinterwordspacing

\bibitem{5596999}
A.~{Horé} and D.~{Ziou}, ``Image quality metrics: Psnr vs. ssim,'' in
  \emph{2010 20th International Conference on Pattern Recognition}, Aug 2010,
  pp. 2366--2369.

\bibitem{1284395}
{Zhou Wang}, A.~C. {Bovik}, H.~R. {Sheikh}, and E.~P. {Simoncelli}, ``Image
  quality assessment: from error visibility to structural similarity,''
  \emph{IEEE Transactions on Image Processing}, vol.~13, no.~4, pp. 600--612,
  April 2004.

\end{thebibliography}
%\end{thebibliography}

% that's all folks
\end{document}